\documentclass[11pt,a4paper]{article}
\usepackage{orcidlink}
\UseRawInputEncoding

\usepackage{amsmath}
\usepackage{amssymb}
\usepackage{amsthm}
\usepackage{comment}
\usepackage{multirow}
\usepackage{booktabs}
\usepackage{authblk} % for author/affiliation formatting
\usepackage[round]{natbib}
\usepackage{float}
\usepackage{bm}
\usepackage{enumerate}
\usepackage{amsfonts}
\usepackage{graphicx}
\usepackage[all]{xy}
\usepackage{tikz}
\usepackage{verbatim}
\usepackage[left=2cm,right=2cm,top=3cm,bottom=2.5cm]{geometry}
\usepackage{hyperref}
\usepackage{caption}
\usepackage{subcaption}
\usepackage{psfrag}
\title{Attention-Based Deep Learning for Early Parkinson's Disease Detection with Tabular Biomedical Data}

\author[1]{Olamide S. Oseni \thanks{Corresponding author: olamide.oseni@aims.ac.rw}\orcidlink{0009-0001-4971-9353}}
\author[2]{Ibraheem O. Obanla \orcidlink{0009-0007-8367-7445}}
\author[3]{Toheeb A. Jimoh \orcidlink{0000-0003-3830-7641}}

\affil[1]{African Institute for Mathematical Sciences, Kigali, Rwanda}
\affil[2]{Department of Data and Information Science, University of Ibadan, Oyo State, Nigeria}
\affil[3]{Department of Computer Science and Information Systems, University of Limerick, Castletroy, V94 T9PX, Limerick, Ireland}

\date{} 
\begin{document}
{\small
\maketitle
}

\noindent%\maketitle
\begin{abstract}
\noindent
Early and accurate detection of Parkinson's disease (PD) remains a critical challenge in medical diagnostics due to the subtlety of early-stage symptoms and the complex, non-linear relationships inherent in biomedical data. Traditional machine learning (ML) models, though widely applied to PD detection, often require extensive feature engineering and fail to capture complex dependencies within and between features. This study investigates the effectiveness of attention-based deep learning models for early PD detection using tabular biomedical data. We conduct a comprehensive empirical comparison of four classification models: Multi-Layer Perceptron (MLP), Gradient Boosting, TabNet, and SAINT. The models were evaluated on a benchmark dataset from the UCI Machine Learning Repository, comprising biomedical voice measurements of PD patients and healthy controls. 

\noindent
Experimental results reveal that the SAINT model consistently outperformed all baseline models across multiple evaluation metrics, achieving a weighted precision of 0.98, a weighted recall of $0.97$, a weighted F1-score of $0.97$, a Matthews Correlation Coefficient (MCC) of $0.9990$, and the highest Area Under the ROC Curve (AUC-ROC). TabNet followed with a weighted precision of 0.96, weighted recall of 0.95, weighted F1-score of 0.95, and MCC of $0.9990$, while MLP achieved a weighted precision of $0.95$, weighted recall of $0.95$, weighted F1-score of $0.95$, and MCC of $0.9995$. Gradient Boosting performed the lowest, with a weighted precision of $0.90$, weighted recall of $0.90$, weighted F1-score of $0.90$, and MCC of $0.7310$. Notably, SAINT's dual attention mechanism enabled superior modelling of feature interactions both within and across the samples, leading to improved predictive performance. 

\noindent
This study establishes the practical feasibility and diagnostic potential of attention-based deep learning architectures, particularly SAINT, for the early detection of Parkinson’s disease. The findings highlight the importance of dynamic feature representation in clinical prediction tasks and advocate for the adoption of advanced attention-based models in healthcare decision-support systems. \\

\end{abstract}
\newpage
\section{Introduction}\label{sec1}
Parkinson's disease (PD) is a progressive neurodegenerative disorder characterized by the degeneration of dopamine-producing neurons, predominantly in the subcortical structure in the brain referred to as the \textit{basal ganglia}. The deterioration of these neurons leads to a cascade of motor dysfunction symptoms, including the characteristic resting tremors, muscle rigidity, bradykinesia, and postural disturbances, among others \citep{kurmi2022_parkinson_ensemble}. PD ranks second among neurodegenerative diseases associated with age, following Alzheimer's disease closely, affecting an estimated range from $6.2$ to $11.77$ million people in recent years, and projected to be over $20$ million people by $2040$ \citep{dorsey2018parkinson_call_to_action, luo2025global}. Moreover, it primarily affects individuals over $60$, with an average onset age of $70$, albeit early-onset cases can occur before $50$ \citep{zhang2022mining}. \\

\noindent
Significant research efforts \citep{Joseph03042023, foote2025_comprehensive} have revealed that there's currently no cure for PD; however, improved medications and therapeutic strategies can manage early symptoms and delay severe disability onset, thereby reducing symptoms, improving the brain function and quality of life, in affected individuals \citep{kilzheimer2019challenge_parkinson_predictive_power}. Consequently, early diagnosis and detection are crucial for optimal management. More explicitly, early diagnosis is essential for the timely initiation of interventions and management strategies \citep{prashanth2018early_parkinson_questionnaire}, thereby enabling the potential detection of the existence or otherwise in susceptible individuals based on symptoms. However, as early detection appears to be a PD management solution, it faces several fundamental challenges. Earliest efforts toward obtaining an accurate diagnosis rely on traditional diagnostic approaches, which involve clinical observations and subjective assessments; thus, they suffer from the lack of definitive biomarkers or clinical tests, particularly in the early stages of the disease development \citep{pahwa2010early_diagnosis}. Subsequently, diagnoses rely on the presence of motor symptoms, which appear late in disease progression when significant neuronal loss has already occurred. Moreover, while non-motor symptoms may even appear years before motor impairments, they lack predictive power \citep{kilzheimer2019challenge_parkinson_predictive_power}. Therefore, researchers have been exploring various modalities for early detection, including finger tapping tests, handwriting analysis, neuroimaging, and voice signals \citep{khanna2020current_parkinson_research}. \\

\noindent
Undoubtedly, the ubiquity of artificial intelligence (AI) techniques in various domains has facilitated the application of more advanced practices for the early detection of PD. The integration of AI computational approaches, particularly machine learning and deep learning algorithms, into the diagnostic process offers promising directions for earlier and more accurate detection \citep{he2024early_ml_dl}. These approaches can detect nuanced and complex patterns in biomedical data that may elude human observation, potentially identifying disease signatures before clinical symptoms manifest. Several studies involving machine learning and deep learning algorithms for early detection of PD have shown promising results \citep{prashanth2018early_parkinson_questionnaire}. More specifically, recent advances in deep learning architectures have introduced increasingly sophisticated models \citep{kumar2024tabnet, pahuja2022deep} for medical diagnosis as they can process complex, multimodal data inputs ranging from neuroimaging and voice recordings to movement patterns and biological markers, potentially revolutionizing early PD diagnosis approaches. However, while previous studies have employed traditional machine and deep learning models for Parkinson's disease detection, these models often fail to model complex inter-feature dependencies or context-specific feature importance explicitly. The advent of attention-based deep learning models \citep{vaswani2017attention} presents a promising frontier for healthcare analytics, with their capability of learning more nuanced patterns, dynamically weighting clinical attributes in the datasets, and improving predictive performance. \\

\noindent
Despite its promising architecture, the potential of an attention-based model like the Self-Attention and Intersample Attention Transformer (SAINT) \citep{saint} in early disease detection tasks, especially in PD detection, remains underexplored. Previous work by \citep{kumar2024tabnet} compared TabNet---another attention-based model---with classical machine learning models for PD detection. However, a comprehensive investigation involving newer architectures like SAINT and other traditional machine learning or deep learning models for early PD remains limited in the literature.  Motivated by these gaps, this study investigates the feasibility and effectiveness of attention-based deep learning models for the early detection of PD. Using the publicly available Parkinson’s dataset from the UCI Machine Learning Repository\footnote{\url{https://archive.ics.uci.edu/}}, we perform a systematic comparative analysis of four models: Multi-Layer Perceptron (MLP), Gradient Boosting, TabNet, and SAINT.
The specific contributions of this study are as follows: 
\begin{itemize}
    \item Presenting a deep learning pipeline for early PD detection and evaluating it on a benchmark biomedical dataset.
    \item Providing a comparative performance analysis of MLP, Gradient Boosting, TabNet, and SAINT, highlighting their strengths and limitations.
    \item Demonstrating the superior performance of SAINT in capturing complex feature interactions within tabular biomedical data.
    \item Discussing the implications of attention-based models like SAINT for future healthcare decision-support systems.
\end{itemize}

\noindent
The findings of this study not only accentuate the potential of deep learning models in medical diagnosis but also provide cogent evidence for adopting attention-based architectures in early disease detection tasks such as PD and other neurodegenerative diseases. 

\section{Related Works}\label{sec2}
Conventional diagnosis of Parkinson's disease has historically relied on clinical assessment of motor symptoms, neurological examination, and response to dopaminergic therapy \citep{conventional_parkinson_disease}. These methods, while established in clinical practice, suffer from significant limitations, including subjectivity, inter-rater variability (IRR), and the inability to detect pre-motor manifestations of the disease. Although the gold standard remains expert neurological assessment, studies indicate that diagnostic accuracy among general practitioners may be as low as $73-80\%$, highlighting the need for more objective and sensitive diagnostic tools.
Neuroimaging techniques such as DaTscan (dopamine transporter scan) have improved diagnostic capabilities, providing visualization of dopaminergic neuronal integrity \citep{he2024early_ml_dl}. However, these approaches remain expensive, not widely accessible, and are typically employed only after clinical symptoms have manifested. The significant gap between disease commencement and clinical diagnosis, which is estimated at $5-10$ years, stresses the urgent need for novel approaches to early detection.
\subsection{Machine Learning Approaches for Early PD Detection}   
The early detection of Parkinson’s disease (PD) has attracted considerable research attention, with machine learning (ML) methods being widely employed due to their ability to learn from complex biomedical data. Traditional ML models have demonstrated promising results in classifying PD using vocal features and other clinical modalities. Among these, Support Vector Machine (SVM) has emerged as a popular choice, achieving classification accuracy as high as $89.74\%$ in one study \citep{SINGH_et_al_2025} and being consistently identified as an efficacious model in another \citep{govindu2023early}. Random Forest has also shown competitive performance, with reported accuracy reaching $91.83\%$ \citep{govindu2023early}. Other classical models, such as Logistic Regression, Decision Trees, and K-Nearest Neighbors (kNN), have also been applied, with kNN achieving up to $95\%$ accuracy in specific comparative studies \citep{CabanillasCarbonell2025Evaluation_zapata}. Ensemble techniques like Adaptive Boosting and Hard Voting have similarly been explored to further improve classification outcomes \citep{K_ensemble_hard_voting}. Collectively, these studies emphasize the potential of ML models in enhancing early PD detection, which is crucial for timely treatment and disease management.

\subsection{Deep Learning Approaches for Early PD Detection}   
Despite their success, traditional ML models often rely heavily on manual feature engineering and may struggle to capture complex, non-linear relationships within biomedical data. To overcome these limitations, recent research has increasingly turned to deep learning (DL) models, which are capable of automatic feature learning and capturing intricate data patterns. Various DL architectures, including Recurrent Neural Networks (RNN), Multilayer Perceptrons (MLP), and Long Short-Term Memory (LSTM) networks, have been applied to PD detection tasks, particularly utilizing voice biomarkers. Notably, LSTM models have reported accuracy levels of up to $99\%$ \citep{Chintalapudi_Cascaded_DL}. Additionally, Convolutional Neural Networks (CNN) have been employed to extract discriminative features from speech signals, hand-drawn images, and time-series signals associated with PD symptoms \citep{biswat_et_al_CNN, Bakkialakshmi_Arulalan_Chinnaraju_Ghosh_Rahat_Saha_2024}, achieving impressive accuracy levels of $93-95\%$. Multi-modal approaches that integrate neuroimaging data with biological and clinical features have also shown promise, with CNN models achieving up to $93.33\%$ accuracy at the feature level and $92.38\%$ at the modal level \citep{pahuja2022deep}.

\subsection{Attention Mechanisms in Deep Learning for Biomedical Data}
While these deep learning approaches represent a significant advancement over traditional ML methods, they often treat all features equally and may not explicitly model complex feature interdependencies or variations across patient records. However, this is a critical need in biomedical data where certain biomarkers may be more influential under specific conditions.
Addressing this challenge, attention-based deep learning models have emerged as powerful tools capable of dynamically weighting feature relevance during prediction. One such model, TabNet \citep{arik2021_tabnet}, has demonstrated its potential in handling tabular data outperforming some classical ML methods. A recent study shows that TabNet achieves an $F1$ score of $83.03\%$ in early PD detection, surpassing support vector machines (SVMs), random forests (RFs), and
decision trees (DTs) \citep{kumar2024tabnet}. TabNet’s sequential attention mechanism enables automatic feature selection and interpretability, offering a promising direction for early PD detection \citep{arik2021_tabnet}.

\subsection{SAINT Model and Advances in Attention-based Models for Tabular Data}
Building on this development, the SAINT model \citep{saint} introduces a more sophisticated architecture that combines both self-attention (intra-sample feature learning) and inter-sample attention (learning from relationships across patient records). SAINT has outperformed conventional models like Gradient Boosting and Random Forest across various benchmark tasks and demonstrated encouraging results in healthcare datasets \citep{gutheil2022saintens, somvanshi2024survey}. Its ability to simultaneously model local and global patterns in tabular data makes it particularly suitable for complex biomedical tasks such as early PD detection.
Recent studies further highlight the effectiveness of attention-based models like TabNet and SAINT in healthcare applications. For example,  SAINTENS, an extension of SAINT, performed comparably to state-of-the-art models in predicting geriatric conditions \citep{gutheil2022saintens}. 
Beyond their superior predictive performance, attention mechanisms have contributed significantly to the advancement of deep learning in biomedical data analysis. They allow models to focus on the most relevant features or data regions, enhancing both accuracy and interpretability \citep{xiao2024exploration, zhang2024memory}. Furthermore, applications of attention-based models in precision medicine have demonstrated their effectiveness in tasks ranging from genomic data analysis and biomarker discovery to disease subtyping and transcription factor binding site prediction \citep{cheng2024attention}.

\subsection{Research Gap and Motivation for this Study}
Despite these advances, there remains a notable gap in the literature regarding the application of attention-based models, particularly SAINT, for the early detection of Parkinson’s disease using structured biomedical data. Existing studies primarily focus on comparing traditional ML and early DL models, with limited exploration of advanced attention mechanisms within this domain. This gap presents a compelling research opportunity to evaluate SAINT’s potential for PD detection and to compare its performance with other prominent models such as MLP, Gradient Boosting, and TabNet.
This study, therefore, aims to fill this gap by conducting a systematic comparative analysis of SAINT against established deep learning and ensemble models for early Parkinson's disease detection. In doing so, it contributes to the growing body of research advocating for attention-based deep learning as a robust and interpretable solution for complex medical classification tasks.

\section{Materials and Methods} \label{sec3}

In this study, we introduce an early diagnosis of PD using a sophisticated attention-based deep learning approach and compare it with other deep learning techniques for performance evaluation. PD is a neurodegenerative disorder that causes symptoms like bradykinesia, rigidity, tremors, and gait disturbances, typically manifesting itself when $50\%$ of the brain neurons are destroyed. The cause is not well understood, although there is a genetic component, environmental factors, and age (over $60$). This section presents the dataset used, along with a comprehensive description of the methods employed in conducting the empirical analysis, which aims to answer the research questions.

\subsection{Dataset}
We utilised the Parkinson's Telemonitoring dataset (ID $174$) from the UCI Machine Learning Repository of the University of California, which classifies patients as PD or not according to speech characteristics. PD detection has two stages: in the training stage, raw data are preprocessed and used for training the model; in the test stage, the model is used for PD detection. \\

\noindent
The aim is to classify whether a person is suffering from Parkinson's disease (binary classification: $1$ for Parkinson's and $0$ for healthy), and to identify a person with that condition early. To detect the early PD, we consider twenty-two features based on information from previous similar studies like \citep{wang2020early, govindu2023early} as provided below: The number of samples used was $195$, and the major features are:

\begin{enumerate}
    \item MDVP:Fo(Hz): Measures the range of the fundamental frequency of speech (in Hz), ranging from $50-500$ Hz, and an extended range may indicate more disordered speech patterns.
    \item MDVP:Fhi(Hz): Represents the peak fundamental frequency (in Hz) of a speech signal ranges from $100-300$ Hz, and a lower peak frequency may indicate stiffness of the vocal cords.
    \item MDVP:Flo(Hz): Shows the lowest fundamental frequency (in Hz) of a speech signal. It is between $50-100$ Hz, and it is dependent on the individual; and lower minimum frequency signifies a lower vocal range, which might be normal in Parkinson's patients.
    \item MDVP:Jitter(\%): Shows the percentage fluctuation (local deviation) in the pitch frequency, usually between $0.1\%$ and $2\%$, and higher jitter shows higher instability in the voice pitch.
    \item MDVP:Jitter(Abs): Provides the absolute value of the pitch variation in frequency, usually expressed in absolute units, e.g., $0-0.5$ Hz, and greater values represent greater pitch instability that is normally associated with Parkinson's due to voice tremors.
    \item MDVP:RAP: Provides the \textit{relative average perturbation} in pitch that quantifies the atypicality in the voice, normally between $0.1$ and $0.6$, and greater RAP indicates greater atypicality in the voice frequency.
    \item MDVP:PPQ: The pitch perturbation quotient measures the total unevenness of the variation of pitch with time, usually, in absolute terms, e.g., $0-0.5$ Hz, and higher values represent greater instability of pitch, which is a typical finding in Parkinson's due to tremor in the voice.
    \item Jitter:DDP: Estimates jitter (pitch frequency change) as the difference of differences. Varies, usually less than $1$, where higher values show more fluctuation in the voice signal, possibly a sign of voice instability typical of Parkinson's.
    \item Shimmer: Measures amplitude perturbation of speech. Typically between $0.1$ and $0.5$. The raised shimmer shows more fluctuation in loudness and may be a sign of motor control problems in the voice.
    \item Shimmer(dB): Quantifies the shimmer in decibels, the variation in voice intensity or loudness. Measured in decibels (dB), usually between $0$ and $3$ dB. The higher the shimmer in dB, the more amplitude changes.
    \item NHR: The Non-Harmonic to Harmonic Ratio, quantifying the ratio of the non-harmonic and harmonic components of speech. Usually between $0$ and $1$. The higher the NHR, the more noise in the voice
    \item HNR: The Harmonic to Noise Ratio, a measurement of the harmonics to noise ratio in the speech signal. Generally between $0$ and $20$ dB. Lower HNR indicates higher levels of noise in the speech signal.
    \item RPDE: Recurrence Period Density Entropy, which measures chaotic activity or complexity in the voice pattern, is generally lower in PD, generally between $0$ and $1$. 
    \item DFA: Detrended Fluctuation Analysis, quantifies the long-term correlation of the speech signal and helps identify small changes in voice dynamics. Typically between $0-1$
    \item Spread1 and Spread2: Helps investigate the variability of Speech.
    \item D2: Correlation dimension for speech signal complexity.
    \item Target Variable: The target label (y) is dichotomous with $0$ indicating the absence of Parkinson's disease and $1$ indicating the presence of Parkinson's disease.
    \item PPE: Also known as Pitch Period Entropy, which specifies the randomness of the pitch period.
\end{enumerate}
\subsection{Exploratory Data Analysis}
\begin{figure}[htbp]
    \centering
    \includegraphics[width=0.75\linewidth]{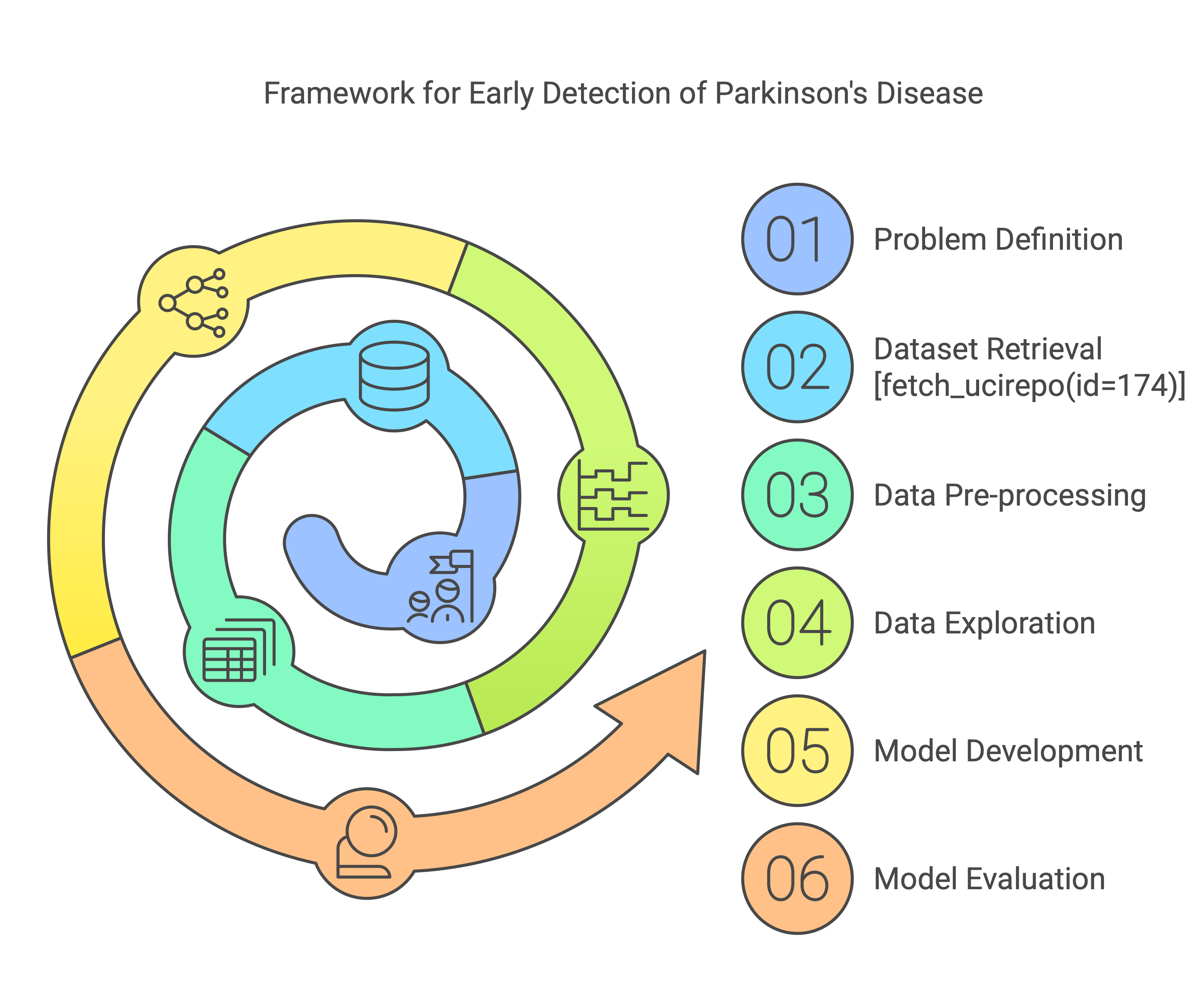}
    \caption{Proposed PD early detection modelling framework}
    \label{fig:framework}
\end{figure}

\begin{figure*}[!ht]
    \centering
    \begin{subfigure}[b]{0.47\textwidth}
        \centering
        \includegraphics[width=\linewidth]{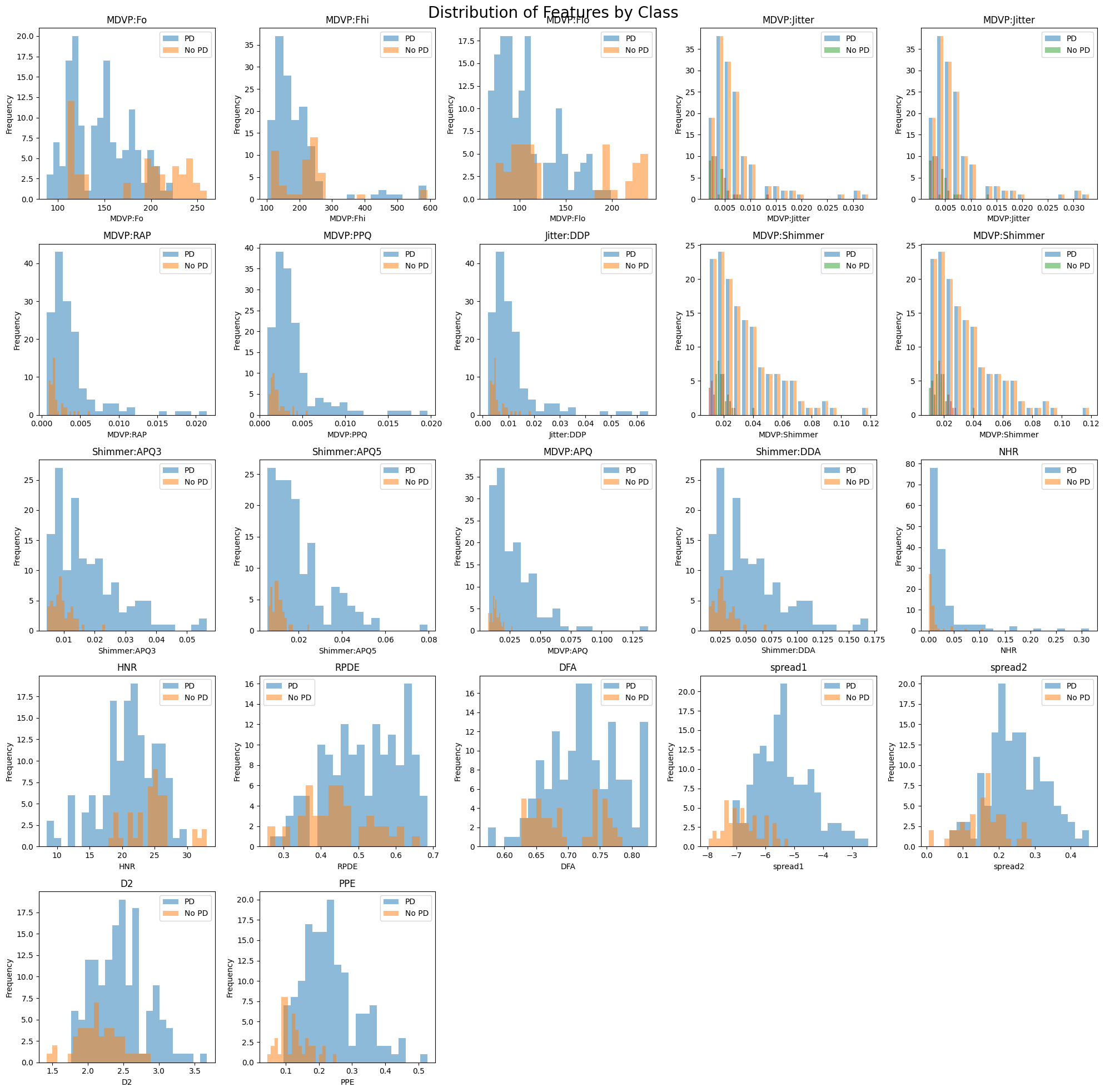}
        \caption{Distribution of features with respect to status (PD or No PD).}
        \label{fig:feature_dist}
    \end{subfigure}
    \hfill
    \begin{subfigure}[b]{0.52\textwidth}
        \centering
        \includegraphics[width=\linewidth]{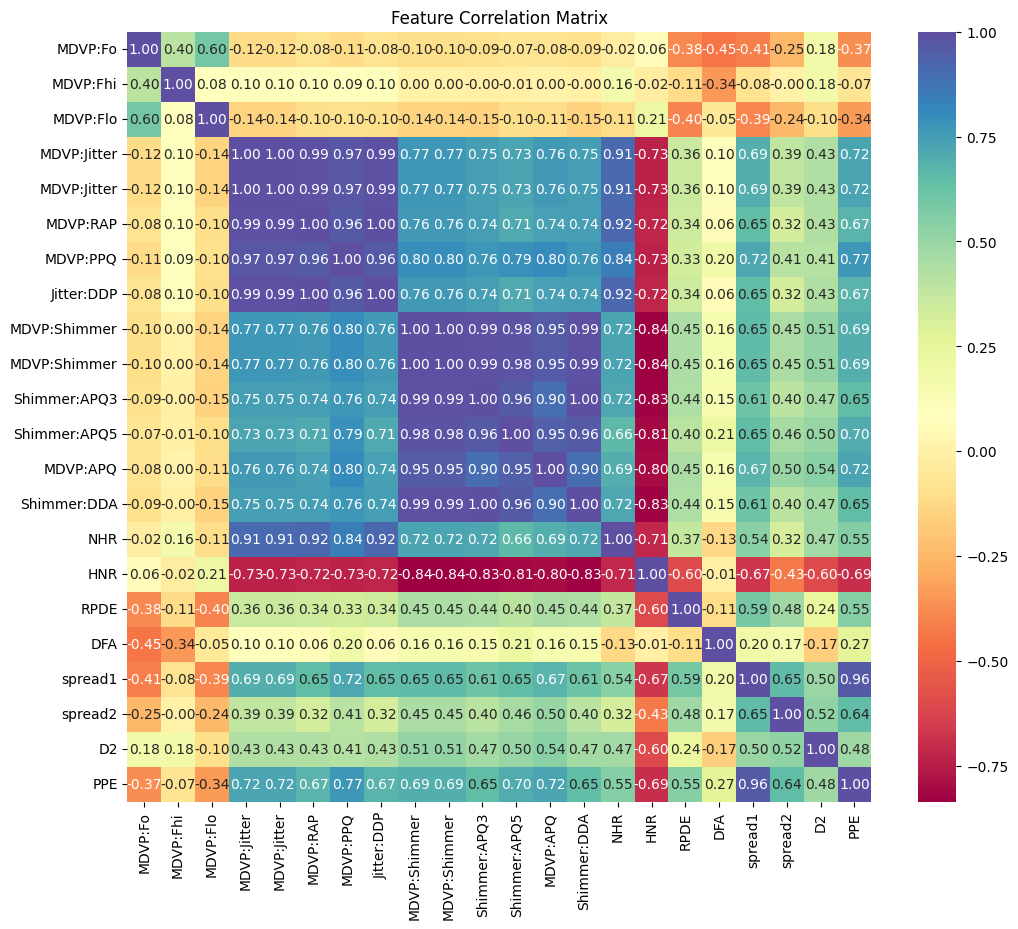}
        \caption{Pairwise Pearson correlation coefficients.}
        \label{fig:correlation}
    \end{subfigure}
    \caption{Exploratory analysis of dataset features.}
    \label{fig:eda}
\end{figure*}

\begin{figure}[h!]
    \centering
    \includegraphics[width=1.0\linewidth]{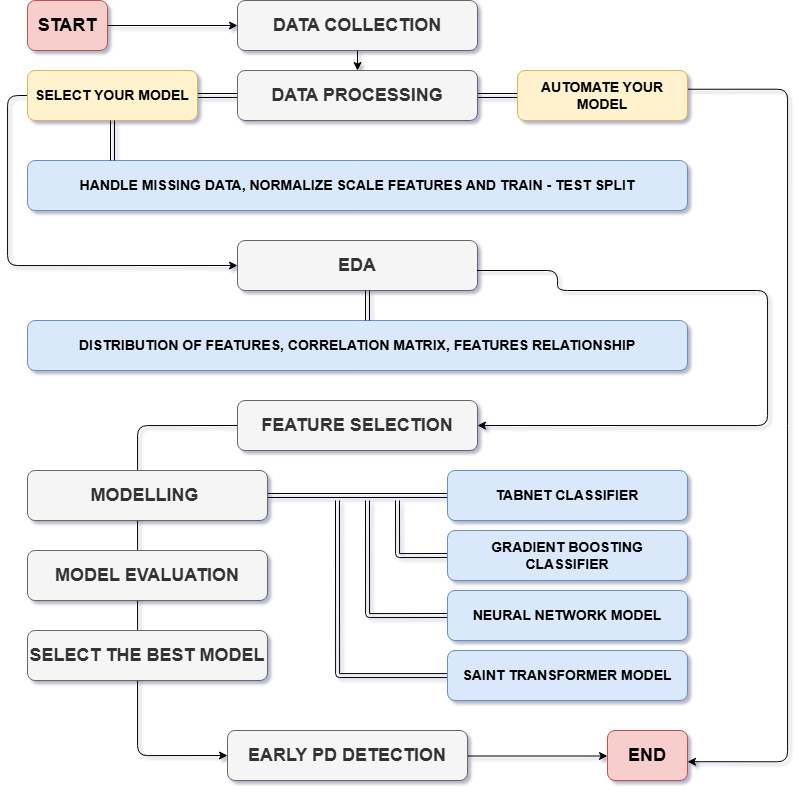}
    \caption{Pipeline for Early Parkinson’s Disease Detection}
    \label{fig:4}
\end{figure}

Figures \ref{fig:framework} and \ref{fig:4} depict the entire modelling pipeline, highlighting the steps required to achieve our goal. Figure \ref{fig:feature_dist} shows histograms of the $22$ features vs. the dependent variable, where $1$ indicates a diagnosis of Parkinson's disease and $0$ indicates no diagnosis. The distribution of the features is left-skewed, which aligns with the finding in \citep{bartl2022longitudinal} of the skewness of health data features, particularly in neurodegenerative diseases. However, \textit{DFA}, \textit{Spread$1$}, and \textit{Spread$2$} features were closer to a normal distribution. Moreover, the Pearson correlation coefficients among all the features are shown in Figure \ref{fig:correlation}. The features Spread $1$ and PPE, as well as \textit{MDVP:Jitter} and \textit{Jitter:DDP}, exhibit a high degree of correlation. Additionally, features such as \textit{MDVP:Jitter}, \textit{MDVP:RAP}, \textit{MDVP:PPQ}, \textit{Jitter:DDP}, and \textit{MDVP:Shimmer} were also strongly correlated with one another. The correlation map displays the pairwise correlation coefficients between the features in the dataset, where blue (cool) colors represent negative correlations and red (warm) colors indicate positive correlations.

% ======== FIGURE 1: Framework ========

% ======== FIGURE 2 & 3: Feature Distribution and Correlation ========

\subsection{Description of Methods}

In this section, we discuss the models used for the detection of Parkinson's disease in its early stages. The approach fuses a number of classification models, including TabNet, Gradient Boosting, Neural Networks, and a custom-designed Transformer-based model called SAINT. We then trained and evaluated the multiple classifiers to identify the most effective model for this task. To begin, the data used to train the model can be defined as:

\begin{equation}
\begin{aligned}
\mathcal{D}_{\text{train}} &= \{ (\mathbf{X}_{\text{train}}, \mathbf{y}_{\text{train}}) \} \\
\mathcal{D}_{\text{test}} &= \{ (\mathbf{X}_{\text{test}}, \mathbf{y}_{\text{test}}) \}
\end{aligned}
\end{equation}

where $\mathbf{X}_{\text{train}} \in \mathbb{R}^{n_{\text{train}} \times d}$ is the input feature matrix for the training set, with $n_{\text{train}}$ samples and $d$ features, and $\mathbf{X}_{\text{test}} \in \mathbb{R}^{n_{\text{test}} \times d}$ is the input feature matrix for the test set, with $n_{\text{test}}$ samples and $d$ features. $\mathbf{y}_{\text{train}} \in \{0, 1\}^{n_{\text{train}}}$ represents the binary class labels for each sample ($0$ for no Parkinson’s disease, $1$ for Parkinson’s disease).

\subsubsection{Techniques}
Deep learning offers a significant advantage in healthcare, allowing more accurate and timely diagnoses \citep{ul2022survey}. \citep{zhang2022mining} stated in his work that newly developed deep learning techniques or deep neural networks are a new set of powerful tools for data classification in PD. 

\begin{figure}[htbp]
    \centering
    \includegraphics[width=1.1\linewidth]{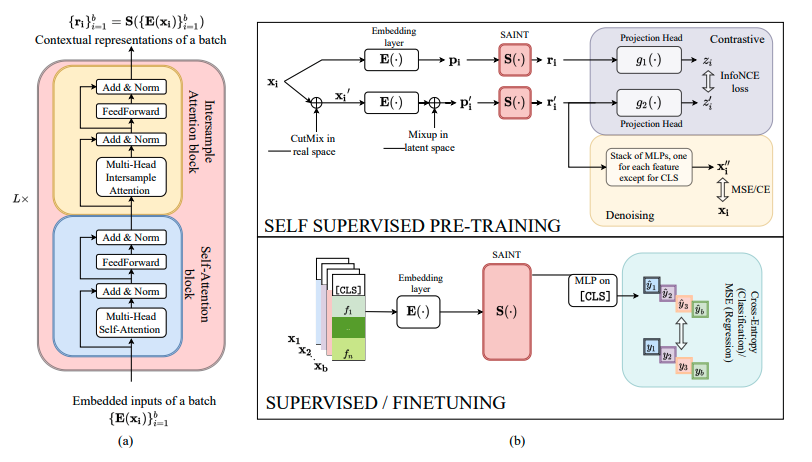}
    \caption{SAINT Architecture, including pre-training and training pipeline introduced by \citep{saint} }
    \label{saint_image}
\end{figure}

\begin{enumerate}
    \item \textbf{TabNet Classifier Learning:}
    TabNet uses attention mechanisms to focus on the most important features of the data and build a decision tree-like structure to enhance performance. Cross-entropy loss minimization was done iteratively through a series of 54 epochs to reduce prediction errors and enhance our model.  This model computes attention coefficients, $\alpha \in \mathbb{R}^{n \times d}$, and applies them to the input data. Let: $\mathbf{X} \in \mathbb{R}^{n \times d}$ be the input feature matrix, where $n$ is the number of samples and $d$ is the number of features. $\alpha \in \mathbb{R}^{n \times d}$ be the attention coefficients learned by the model. $\hat{\mathbf{X}} \in \mathbb{R}^{n \times d}$ be the weighted input features after applying attention.
The model applies the attention coefficients to the input data:

\begin{equation}
\hat{\mathbf{X}} = \text{Softmax}(\alpha) \cdot \mathbf{X}
\end{equation}

Then, cross-entropy loss was used to evaluate the model's accuracy in classifying whether a patient has Parkinson's disease. Let:

\begin{equation}
L = - \sum_{i=1}^{n} \left[ y_i \log(\hat{y}_i) + (1 - y_i) \log(1 - \hat{y}_i) \right]
\end{equation}

Where $y_i \in \{0, 1\}$ be the binary class label for sample $i$, where $y_i = 1$ indicates Parkinson's disease and $y_i = 0$ indicates no Parkinson's disease. $\hat{y}_i$ be the predicted probability for sample $i$ and$n$ be the number of samples in the data set.

\item \textbf{Gradient Boosting Classifier:}
\citep{anisha2020early} proposed Gradient Boosting in their research on PD detection, which they consider as the building of a sequence of models within an ensemble, such that each successive model attempts to compensate for the errors of the previous model. Each model is learned on the residual errors of the last model, and the final prediction is the sum of the predictions of all the individual models.

The final prediction is the sum of all previous models:

\begin{equation}
F(\mathbf{X}) = \sum_{m=1}^{M} f_m(\mathbf{X})
\end{equation}

where $F(\mathbf{X})$ is the final prediction, and $f_m(\mathbf{X})$ represents the prediction from the $m$-th model, with $M$ being the total number of models. At each stage, the model minimizes residuals by fitting the next model $f_m(\mathbf{X})$ to the residuals:

\begin{equation}
L = \frac{1}{n} \sum_{i=1}^{n} \left( y_i - \hat{y}_i \right)^2
\end{equation}

where $n$ is the number of samples, $y_i$ is the true label for the $i$-th sample, and $\hat{y}_i$ is the predicted label for the $i$-th sample. The residuals are updated through gradient descent, guiding the model to progressively minimize the error:

\begin{equation}
\hat{y}_i^{(m+1)} = \hat{y}_i^{(m)} + \eta \cdot f_m(\mathbf{X}_i)
\end{equation}

where $\hat{y}_i^{(m+1)}$ is the updated prediction for the $i$-th sample at the $(m+1)$-th stage, $\hat{y}_i^{(m)}$ is the prediction at the $m$-th stage, and $\eta$ is the learning rate. The function $f_m(\mathbf{X}_i)$ is the prediction from the $m$-th model for the $i$-th sample.

\item \textbf{Neural Network Classifier (MLP):}
Then, the classification was performed by a Multi-Layer Perceptron (MLP). There are multiple layers of neurons in the neural network, each of which transforms the input through weighted sums and nonlinear activation functions. It aims to map clinical data features to a binary outcome, reflecting whether a patient has Parkinson's disease.

Each layer performs a weighted sum of the input followed by a non-linear activation function, such as ReLU. The transformation for the \(l\)-th layer is:

\begin{equation}
\mathbf{h}^{(l)} = \sigma \left( \mathbf{W}^{(l)} \mathbf{X} + \mathbf{b}^{(l)} \right)
\end{equation}

where \( \mathbf{h}^{(l)} \) is the output of the \(l\)-th layer, \( \mathbf{W}^{(l)} \) is the weight matrix for the layer, \( \mathbf{b}^{(l)} \) is the bias vector, \( \mathbf{X} \) is the input to the layer, and \( \sigma \) is the activation function (e.g., ReLU) and the final layer computes the probability of class membership using the sigmoid function:

\begin{equation}
\hat{y} = \sigma \left( \mathbf{W}^{(L)} \mathbf{h}^{(L-1)} + \mathbf{b}^{(L)} \right)
\end{equation}

where \( \hat{y} \) is the predicted probability, \( \mathbf{h}^{(L-1)} \) is the output from the previous layer (layer \( L-1 \)), and \( \mathbf{W}^{(L)} \) and \( \mathbf{b}^{(L)} \) are the weights and biases for the final layer. We then used binary cross-entropy loss to evaluate the classification error:

\begin{equation}
L = - \sum_{i=1}^{n} \left[ y_i \log(\hat{y}_i) + (1 - y_i) \log(1 - \hat{y}_i) \right]
\end{equation}

where \(n\) is the number of samples, \(y_i\) is the true label for sample \(i\), and \( \hat{y}_i \) is the predicted probability for sample \(i\).

\item \textbf{SAINT Model:}
Finally, we used the SAINT architecture, depicted in Figure \ref{saint_image}, which is based on the Transformer architecture. Originally designed for sequence data, SAINT adapts the self-attention mechanism for tabular data, modelling dependencies between features that may be essential for disease classification. The input features are first embedded into a higher-dimensional space:

\begin{equation}
\mathbf{X}_{\text{emb}} = \mathbf{W}_{\text{emb}} \mathbf{X} + \mathbf{b}_{\text{emb}}
\end{equation}
Where
 \( \mathbf{X}_{\text{emb}} \in \mathbb{R}^{n \times d_{\text{emb}}} \) is the embedded input feature matrix,
 \( \mathbf{W}_{\text{emb}} \in \mathbb{R}^{d_{\text{emb}} \times d} \) is the embedding weight matrix,
 \( \mathbf{b}_{\text{emb}} \in \mathbb{R}^{d_{\text{emb}}} \) is the embedding bias vector,
\( \mathbf{X} \in \mathbb{R}^{n \times d} \) is the input feature matrix.

\begin{equation}
\alpha_{ij} = \frac{\exp(\mathbf{Q}_i \mathbf{K}_j^T)}{\sum_{k} \exp(\mathbf{Q}_i \mathbf{K}_k^T)}
\end{equation}

where:
 \( \alpha_{ij} \) represents the attention score between feature \( i \) and feature \( j \),
 \( \mathbf{Q}_i \in \mathbb{R}^{d_{\text{emb}}} \) is the query vector for feature \( i \),
 \( \mathbf{K}_j \in \mathbb{R}^{d_{\text{emb}}} \) is the key vector for feature \( j \),
 \( d_{\text{emb}} \) is the embedding dimension.  After passing through the transformer layers, the final output was computed using the following.

\begin{equation}
\hat{y} = \sigma \left( \mathbf{W}_{\text{out}} \mathbf{h}_{\text{final}} + \mathbf{b}_{\text{out}} \right)
\end{equation}

where:
 \( \hat{y} \in [0,1] \) is the predicted probability for the class,
 \( \mathbf{W}_{\text{out}} \in \mathbb{R}^{d_{\text{final}} \times 1} \) is the output weight matrix,
 \( \mathbf{h}_{\text{final}} \in \mathbb{R}^{d_{\text{final}}} \) is the output from the final transformer layer,
 \( \mathbf{b}_{\text{out}} \in \mathbb{R} \) is the output bias,
 \( \sigma \) is the sigmoid activation function.
\end{enumerate}

\subsubsection{Model Evaluation}

To compare the performance of the mentioned algorithms in early Parkinson's disease detection, the data was split randomly, with $80\%$ reserved for training purposes and $20\%$ reserved for testing. For ensuring reproducibility, the random state of $42$ was considered. The same proportion of people who had PD and did not have PD was kept by stratified sampling as in the original data. The performance metrics of the test, including accuracy, AUC-ROC, sensitivity, F1 score, recall, and the area under the ROC curve, were computed and used for evaluation. For the deep learning models, the training was done with $100$ epochs, a batch size of 256, and a virtual batch size of $128$, specifically for the TabNet model.\\

\noindent
Generally, to measure the effectiveness of the models for early Parkinson’s disease detection, several standard classification metrics were utilized. This includes the four elements of the confusion matrix for PD binary classification, including the Area Under the ROC Curve (AUC) and the Matthews Correlation Coefficient (MCC), which were also calculated.  The description, alongside their mathematical representations, is given in the next sections.

\subsubsection{Evaluation Metrics}
Given a binary classification task with confusion matrix elements as True Positives $(\widehat{\text{TP}})$, True Negatives $(\widehat{\text{TN}})$, False Positives $(\widehat{\text{FP}})$, and False Negatives $(\widehat{\text{FN}})$. 

    \begin{equation}
        \text{Accuracy} = \frac{\widehat{\text{TP}} + \widehat{\text{TN}}}{\widehat{\text{TP}} + \widehat{\text{FP}} + \widehat{\text{TN}} + \widehat{\text{FN}}}
    \end{equation}

    \begin{equation}
        \text{Sensitivity} = \frac{\widehat{\text{TP}}}{\widehat{\text{TP}} + \widehat{\text{FN}}}
    \end{equation}

    \begin{equation}
        \text{Specificity} = \frac{\widehat{\text{TN}}}{\widehat{\text{TN}} + \widehat{\text{FP}}}
    \end{equation}

    \begin{equation}
        \text{Precision} = \frac{\widehat{\text{TP}}}{\widehat{\text{TP}} + \widehat{\text{FP}}}
    \end{equation}

    \begin{equation}
        F1 = 2 \times \frac{\text{Precision} \times \text{Recall}}{\text{Precision} + \text{Recall}}
    \end{equation}  \\

    \noindent
The Accuracy measures the proportion of correct predictions the model makes, with a higher accuracy value indicating better overall performance. Sensitivity indicates the model's ability to identify Parkinson’s patients correctly; Recall is equivalent to sensitivity in binary classification, and Specificity measures how well the model identifies healthy individuals. 

\begin{enumerate}
    \item \textbf{Weighted Average Metric:} 

 To account for class imbalance, the weighted average of metric $M$, such as Precision, Recall, or F1-score, is computed as:
\begin{equation}
M_{weighted} = \sum_{i=1}^{k} w_i \times M_i
\end{equation} \\

where:
\begin{itemize}
    \item $k$ is the number of classes,
    \item $M_i$ is the metric score for class $i$,
    \item $w_i = \frac{n_i}{N}$, with $n_i$ being the number of samples in class $i$ and $N$ the total number of samples.
\end{itemize}

\vspace{0.5em}

\item \textbf{Area Under the ROC Curve (AUC-ROC)}: \\
AUC-ROC evaluates the model's ability to distinguish between classes:
\begin{equation}
\text{AUC-ROC} = \int_{0}^{1} TPR(FPR) \, dFPR
\end{equation}

where $TPR$ is the True Positive Rate and $FPR$ is the False Positive Rate.

\vspace{0.5em}

\item \textbf{Matthews Correlation Coefficient (MCC):}\\
MCC is a robust metric, particularly effective for imbalanced datasets:
\begin{equation}
MCC = \frac{\left(\widehat{\text{TP}} \times \widehat{\text{TN}}\right) - \left(\widehat{\text{FP}} \times \widehat{\text{FN}}\right)}{\sqrt{\left(\widehat{\text{TP}} + \widehat{\text{FP}}\right)\left(\widehat{\text{TP}} + \widehat{\text{FN}}\right)\left(\widehat{\text{TN}} + \widehat{\text{FP}}\right)\left(\widehat{\text{TN}} + \widehat{\text{FN}}\right)}}
\end{equation}

MCC ranges from $-1$ to $1$, where:
\begin{itemize}
    \item $1$ indicates perfect prediction,
    \item $0$ indicates random prediction,
    \item $-1$ indicates total disagreement between the prediction and ground truth.
\end{itemize}

\end{enumerate}

\section{Results and Interpretations}

In this section, we present an evaluation and comparison of the performances of the four models that include our newly introduced SAINT model, as well as TabNet, MLP, and gradient boosting (GBM). Performance assessment is performed using multiple complementary metrics that include precision, recall, F1 score, ROC analysis, and MCC, which have been previously illustrated in Section 3.2.2. In addition, weighted averages of precision, recall, and F1 score were calculated and reported to account for any class imbalances in the data, helping us to better understand how well each model works in different groups of patients. \\

% ======== FIGURE 1: Performance Metrics ========
\begin{figure*}[!ht]
    \centering
    % Row 1
    \begin{subfigure}[b]{0.45\textwidth}
        \centering
        \includegraphics[width=\linewidth]{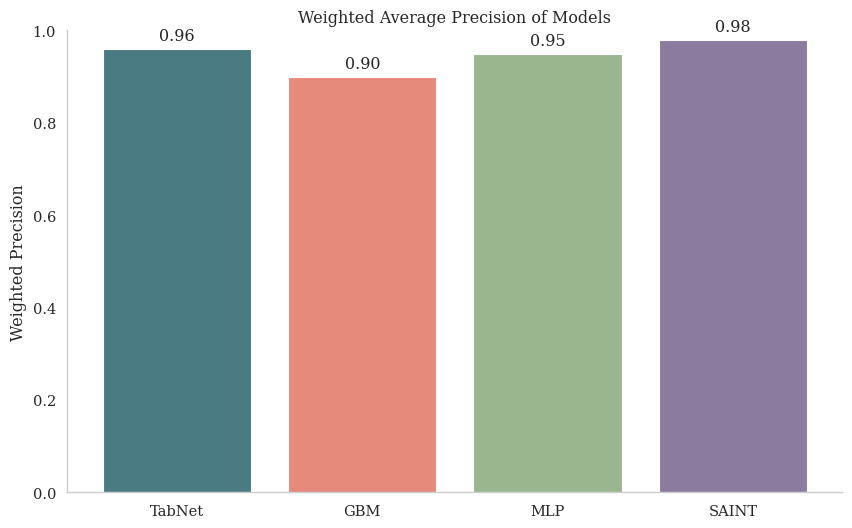}
        \caption{Weighted Precision}
        \label{fig:precision}
    \end{subfigure}
    \hfill
    \begin{subfigure}[b]{0.45\textwidth}
        \centering
        \includegraphics[width=\linewidth]{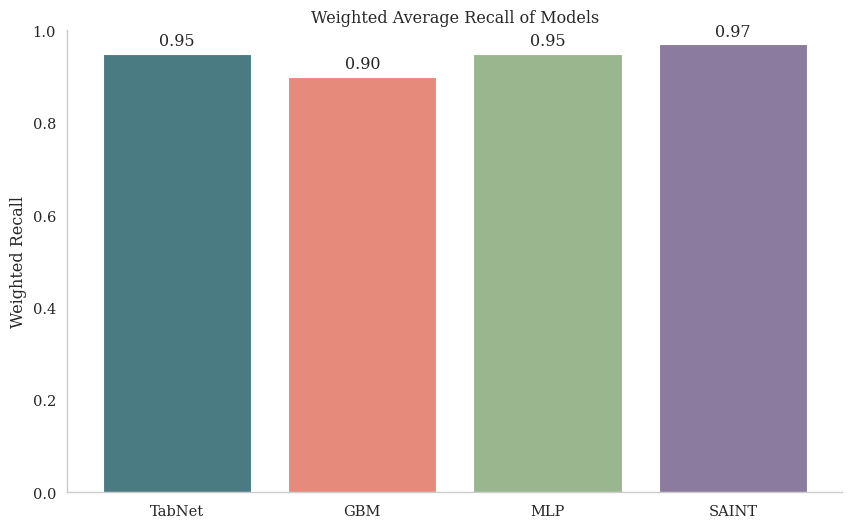}
        \caption{Weighted Recall}
        \label{fig:recall}
    \end{subfigure}

    \vspace{0.4cm}

    % Row 2
    \begin{subfigure}[b]{0.45\textwidth}
        \centering
        \includegraphics[width=\linewidth]{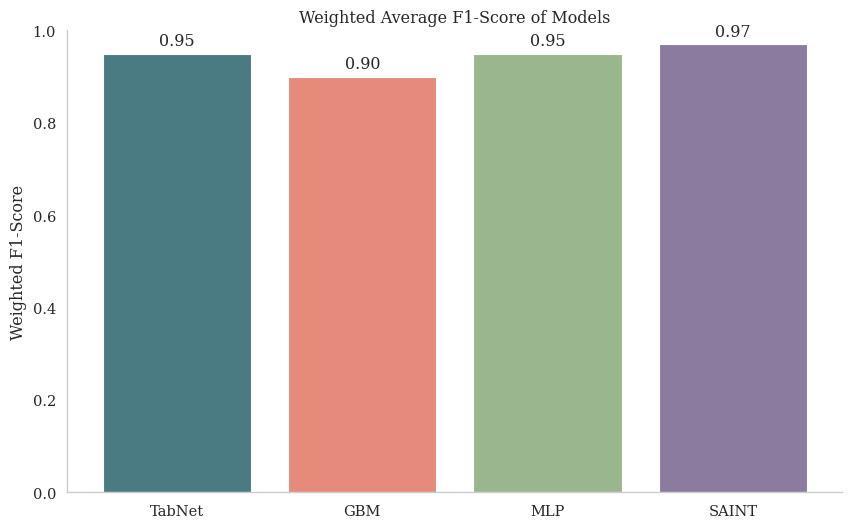}
        \caption{Weighted F1-Score}
        \label{fig:F1}
    \end{subfigure}
    \hfill
    \begin{subfigure}[b]{0.45\textwidth}
        \centering
        \includegraphics[width=\linewidth]{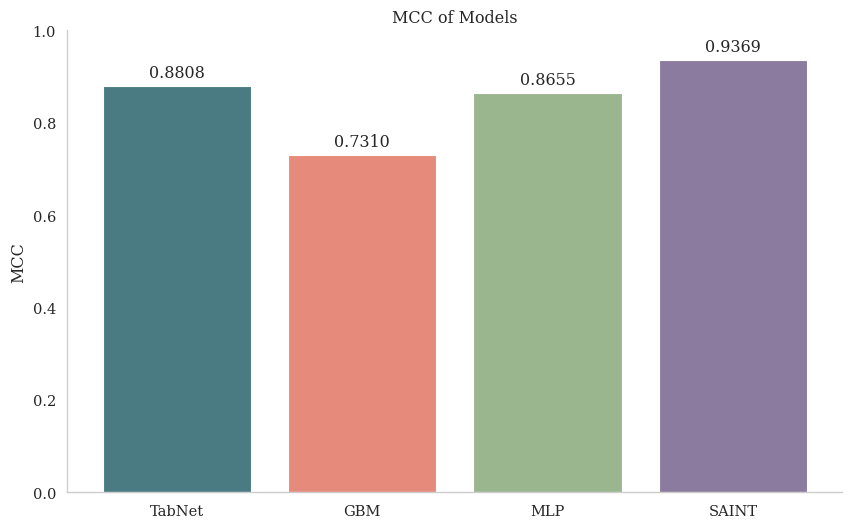}
        \caption{MCC}
        \label{fig:MCC}
    \end{subfigure}

    \caption{Comparison of performance metrics for different models.}
    \label{fig:performance_metrics}
\end{figure*}

% ======== FIGURE 2: Confusion Matrices ========
\begin{figure*}[!ht]
    \centering
    % Row 1
    \begin{subfigure}[b]{0.45\textwidth}
        \centering
        \includegraphics[width=\linewidth]{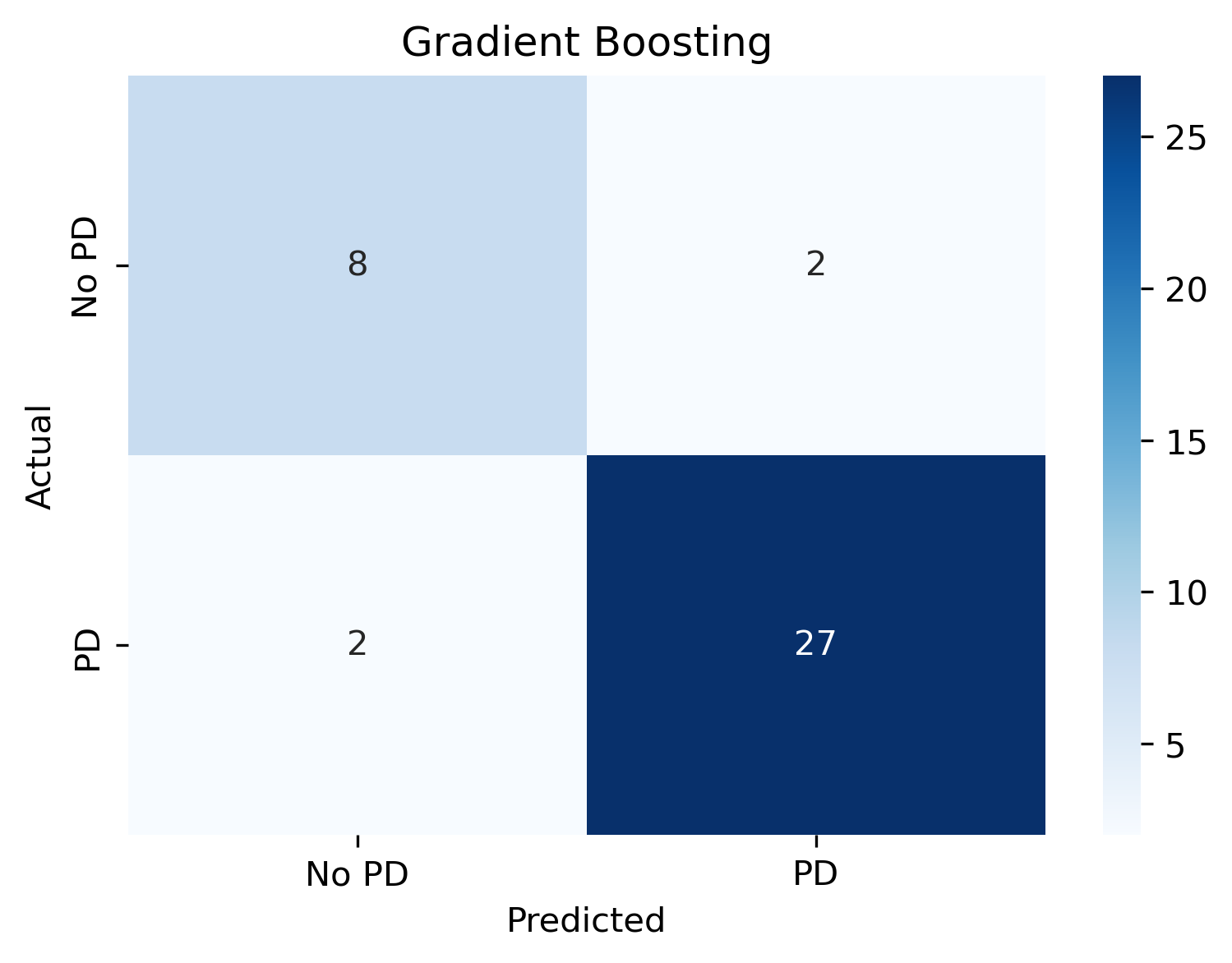}
        \caption{GBM}
        \label{fig:GBM}
    \end{subfigure}
    \hfill
    \begin{subfigure}[b]{0.45\textwidth}
        \centering
        \includegraphics[width=\linewidth]{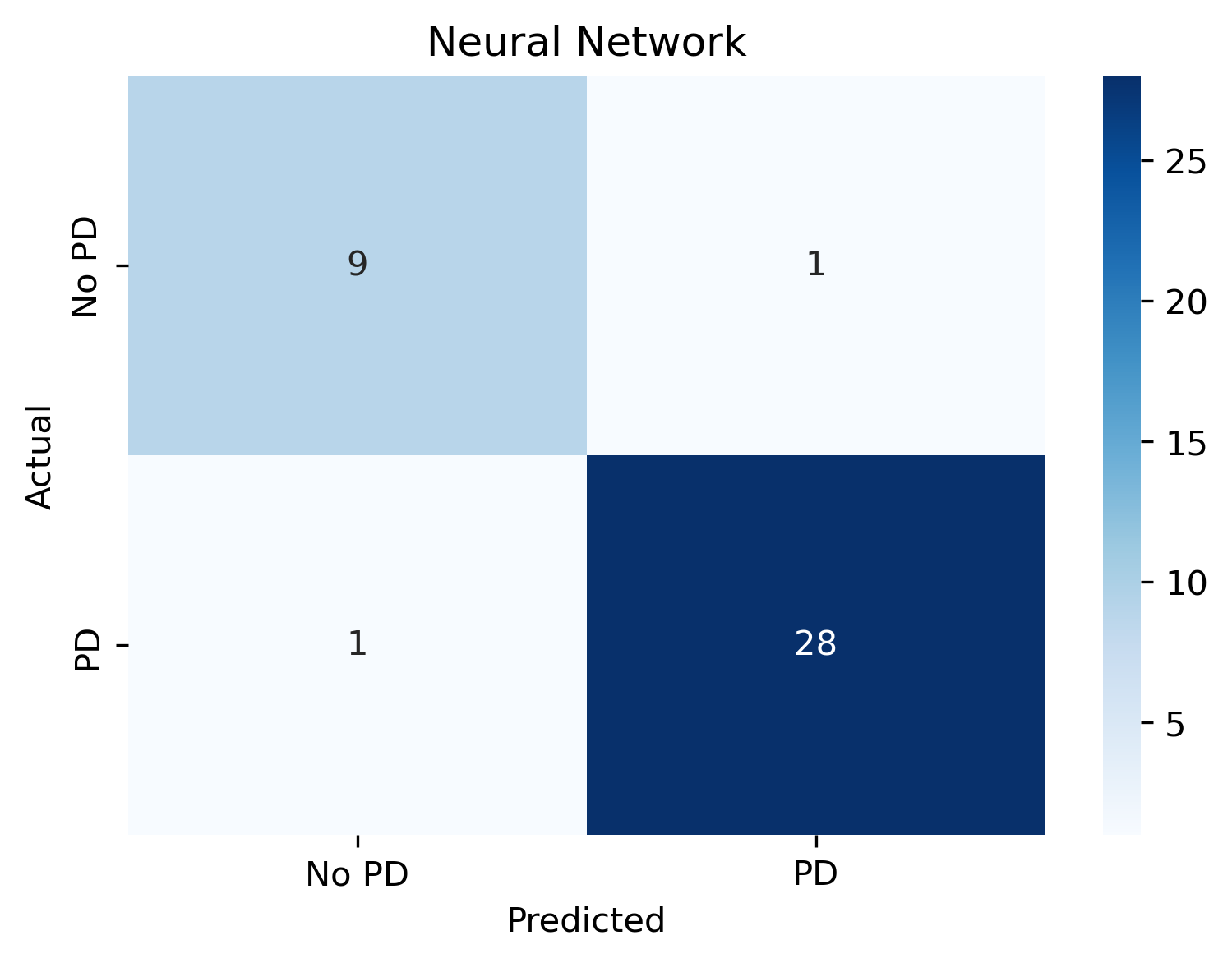}
        \caption{MLP}
        \label{fig:MLP}
    \end{subfigure}

    \vspace{0.4cm}

    % Row 2
    \begin{subfigure}[b]{0.45\textwidth}
        \centering
        \includegraphics[width=\linewidth]{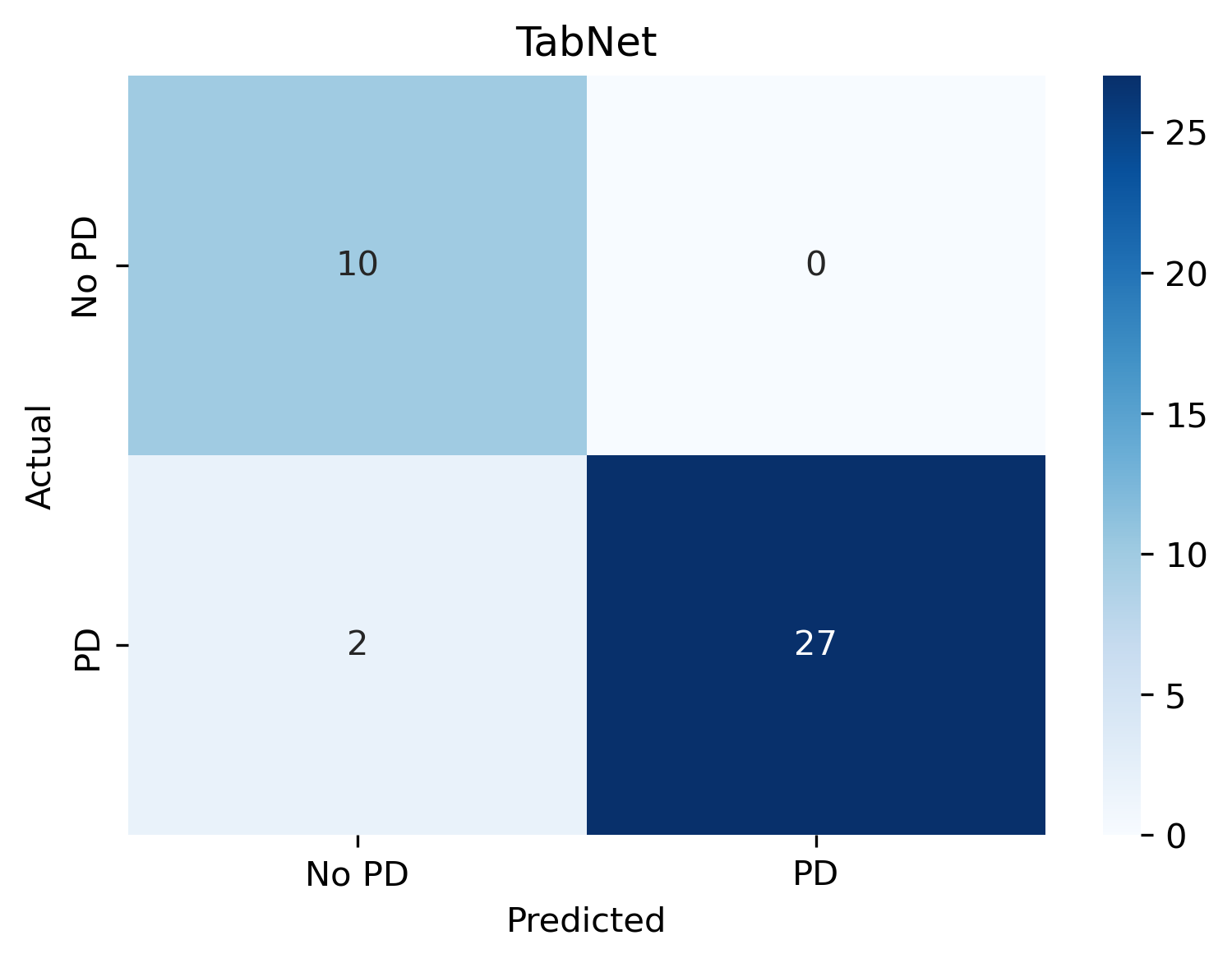}
        \caption{TabNet}
        \label{fig:TabNet}
    \end{subfigure}
    \hfill
    \begin{subfigure}[b]{0.45\textwidth}
        \centering
        \includegraphics[width=\linewidth]{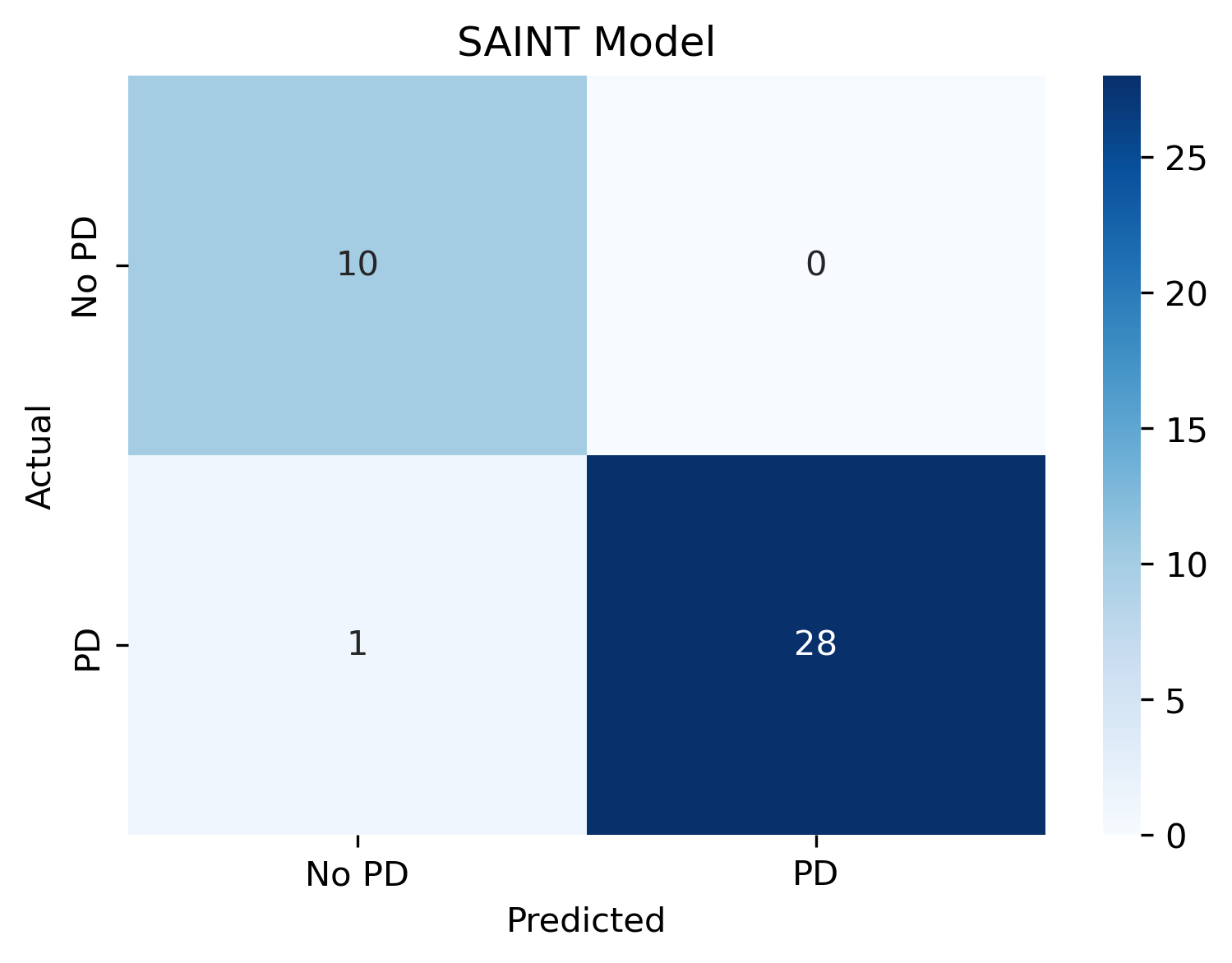}
        \caption{SAINT}
        \label{fig:SAINT}
    \end{subfigure}

    \caption{Confusion matrices for the four models.}
    \label{fig:confusion_matrices}
\end{figure*} 

\noindent
The weighted averages of precision, recall, and $F1$ score, as well as MCC, are shown in Figure~\ref{fig:performance_metrics}. The weighted precision metrics indicate how well the models can avoid false-positive diagnoses. SAINT demonstrated the highest precision $(0.98)$, indicating that when it predicted a positive Parkinson's case, it was correct $98\%$ of the time. TabNet followed closely $(0.96)$, with MLP $(0.95)$ and GBM $(0.90)$ showing progressively lower precision values. This gradient in precision performance is particularly noteworthy in the clinical context, where false-positive diagnoses can lead to unnecessary interventions and psychological distress. 

\noindent
The weighted recall values indicate how well the models can identify all positive cases. Again, SAINT outperformed other models with a recall of $0.97$, successfully identifying $97\%$ of actual Parkinson's cases. TabNet and MLP achieved recall values of $0.95$, while GBM lagged with $0.90$. These recall metrics are especially important for early detection applications, as missed diagnoses (false negatives) can delay critical early intervention for Parkinson's disease.\\

\noindent
The weighted average F1 score indicates the overall effectiveness of the models in balancing precision and recall. SAINT achieved the highest F1 score of $0.97$, indicating superior overall performance in correctly identifying both Parkinson's cases and controls while minimizing false classifications. TabNet and MLP both attained $F1$ scores of $0.95$. GBM achieved the lowest $F1$ score at $0.90$, suggesting relatively reduced effectiveness in balancing false positives and false negatives compared to the other models.\\

\noindent
The MCC values provide crucial information about the models' performance regarding potential class imbalance, which is relevant for medical diagnostics. SAINT exhibited the highest MCC value of $0.9369$, indicating an exceptional correlation between predicted and actual classifications. TabNet and MLP followed with respectable values of $0.8808$ and $0.8655$, respectively. GBM showed the lowest MCC at $0.7310$, suggesting less reliable binary classification performance, potentially due to increased sensitivity to class distribution challenges in the dataset. \\

\noindent
The confusion matrices give us a closer look at how each model performed on an individual case basis, revealing important details that summary metrics might not fully capture.
The confusion matrix for the GBM, shown in Figure~\ref{fig:GBM}, indicates the lowest accuracy among the four models. It correctly classified 8 cases as No PD, and $2$ cases were falsely predicted as PD. For PD cases, $27$ were correctly identified, while $2$ were missed as No PD. This resulted in $2$ false positives and $2$ false negatives, which contributed to a recall and precision of $0.90$ each. \\

\noindent
The confusion matrix for the MLP model shown in Figure~\ref{fig:MLP} provides evidence of a balanced performance between PD and No PD. It correctly identified $9$ cases as No PD, and $1$ case was incorrectly predicted as PD. For PD cases, $28$ were accurately classified, and only $1$ case was missed as No PD. This translates to a recall and precision close to $0.95$, as reported earlier. \\

\noindent
The confusion matrix for the TabNet model shown in Figure~\ref{fig:TabNet} shows a solid performance, as it correctly classified $10$ cases as No PD, with no false positives, which means that no healthy individuals were incorrectly labeled as having PD. However, in the case of PD, $27$ were accurately classified while $2$ cases were wrongly identified as No PD. This is consistent with the reported recall of $0.95$ and precision of $0.96$. \\

% ======== FIGURE 3: ROC Curve ========
\begin{figure}[htbp]
    \centering
    \includegraphics[width=0.78\linewidth]{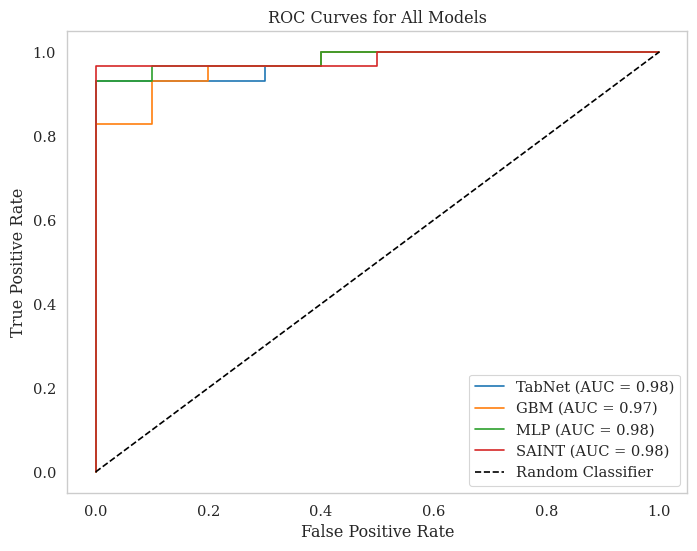}
    \caption{ROC Curve for the models.}
    \label{fig:ROC}
\end{figure}

\noindent
The confusion matrix of the SAINT model, which is the overall best-performing model, is shown in Figure ~\ref{fig:SAINT}. The matrix indicates that out of the total samples, 10 cases were correctly predicted as No PD, with no instances of false positives where non-PD cases were mistakenly identified as PD. In terms of PD cases, 28 were correctly identified, and only 1 case was incorrectly predicted as No PD. This result shows that the proposed SAINT model accurately distinguishes between healthy individuals and those with PD, with a minimal number of missed diagnoses. The low false negative rate aligns with the reported recall of 0.97 and precision of 0.98, and this supports the dual attention mechanism proposed by our study to effectively capture complex feature interactions in the tabular biomedical data. 

\noindent
Furthermore, Figure~\ref{fig:ROC} displays the ROC curve, which illustrates how well each model distinguishes between PD and non-PD cases. The curve plots the True Positive Rate against the False Positive Rate for all the models. As observed, the SAINT model achieved the highest AUC, which shows that it is the strongest in identifying PD cases while minimizing false positives. The TabNet model follows with a comparable result, while MLP and GBM show lower AUC scores, which are good results, but have weaker classification performance compared to the SAINT model. These results provide further evidence that our proposed model, SAINT, offers the most reliable detection of PD in this study.

\section{Discussion} \label{sec5}

In this study, we examined the use of attention-based deep learning models for the early detection of Parkinson’s disease (PD) using biomedical voice data. Our results demonstrated that SAINT, which is an attention-based model that incorporates both self-attention and inter-sample attention mechanisms, produced the most reliable classification performance when compared to TabNet, MLP, and GBM. \\

\noindent
Attention-based models have emerged as a response to the limitations of traditional approaches when it comes to capturing complex feature relationships in biomedical data. This can be seen in the study by \citep{kumar2024tabnet}, where TabNet outperformed traditional models in detecting PD from biomedical data, as a result of its ability to select and interpret features automatically. While this represented a significant step forward, our findings show that SAINT advances this approach further by integrating both intra- and inter-sample attention. This dual attention mechanism appears to allow SAINT to account not only for relationships in individual patient records but also for patterns that emerge across different samples. In addition, this feature selection in SAINT is important when dealing with medical data where subtle variations between patients have diagnostic significance.\\

\noindent
The results support the high performance of SAINT, but it is important to consider the context of the results. Although the dataset applied in the current study is the same as commonly found in the literature, it is rather small and concentrates on features based solely on voice. Since the disease has a broad impact on motor and nonmotor activities, the use of voice data, which is convenient and not invasive, may present an incomplete picture of the expression of the disease. In addition to that, the high potential of representation learning of deep learning models, such as SAINT, can be easily overfitted in working with a small set of data, even under the conditions that suitable regularization strategies can be implemented. Therefore, to improve future research, it would be advantageous to expand the size and variety of data used and include multimodal clinic data.\\

\noindent
From a clinical view, the SAINT and other attention-based models are practical in making healthcare decisions. Their capability of making superior biomedical data processing with structured information is one of the pathways to realizing better and timely PD diagnosis. It should be noted, however, that the nature of these models may be difficult to interpret. The decisions made in the medical area need to be not only accurate in prediction but also easily expounded on and articulated both to the clinicians and the patients. Though the concept of SAINT brings about improvements in feature representation, more studies are necessary to evaluate its interpretability as applied to clinical applications.\\

\noindent
Our study provides evidence for the expanding body of literature advocating attention-based deep learning in medical diagnostics. The findings support the prospect of using SAINT to increase early PD detection, although the conclusions point to the necessity to validate the solution on larger and multi-modal datasets and the need to plan how such models can be used in clinical practice.\\

\section{Conclusions} \label{sec6}

In conclusion, our study has contributed to the growing practical application of attention-based deep learning models in clinical settings such as the early detection of Parkinson's disease. A systematic comparison with other models, such as TabNet, MLP, and GBM, reveals that SAINT offers better classification accuracy, which shows that the model applies to the early-stage PD detection problem. These results contribute to prior findings by \citep{kumar2024tabnet}, by showing that more advanced attention mechanisms can enhance predictive accuracy in structured medical data. \\

\noindent
However, it is important to note that these results are based on a specific dataset that focuses on voice recordings. The dataset also involved a relatively small sample, which limits how far the findings can be generalized to the wider population. The dataset was limited to speech-based acoustic features (e.g., jitter, shimmer, fundamental frequency measures), which are highly informative but do not capture the full clinical spectrum of Parkinson’s disease. Future research should develop larger and more diverse datasets that include demographic and clinical metadata (such as age, sex, and medication use), longitudinal data to track disease progression, which could be time sensitive, as there is a current limitation in the availability of such valuable datasets.\\

\noindent
Despite these limitations, the study demonstrates the potential of attention-based models to support early detection of PD. With further testing and refinement, such models could contribute meaningfully to clinical decision-making processes, especially in the presence of a limited dataset, considering the scarcity of relevant data for carrying out contemporary research in this field. \\

\noindent
\textbf{Declaration of Conflict of Interest}\\
The authors certify that no conflict of interest exists among them in respect of this publication. \\

\noindent
\textbf{Acknowledgements}\\
The authors appreciate the reviewers and the editor for their efforts towards growth and fairness in research. \\

\noindent
\textbf{Data Availability}\\
The research dataset is the Parkinson's Telemonitoring dataset (ID 174) from the UCI Machine Learning Repository\footnote{\url{https://archive.ics.uci.edu/}} of the University of California. 

\newpage
\renewcommand{\bibname}{references}

%\nocitep{*}
\bibliographystyle{plainnat}
\bibliography{references}

\begin{thebibliography}{33}
\providecommand{\natexlab}[1]{#1}
\providecommand{\url}[1]{\texttt{#1}}
\expandafter\ifx\csname urlstyle\endcsname\relax
  \providecommand{\doi}[1]{doi: #1}\else
  \providecommand{\doi}{doi: \begingroup \urlstyle{rm}\Url}\fi

\bibitem[Anisha and Arulanand(2020)]{anisha2020early}
CD~Anisha and N~Arulanand.
\newblock Early prediction of parkinson's disease (pd) using ensemble classifiers.
\newblock In \emph{2020 International Conference on Innovative Trends in Information Technology (ICITIIT)}, pages 1--6. IEEE, 2020.

\bibitem[Arik and Pfister(2021)]{arik2021_tabnet}
Sercan~{\"O} Arik and Tomas Pfister.
\newblock Tabnet: Attentive interpretable tabular learning.
\newblock In \emph{Proceedings of the AAAI conference on artificial intelligence}, volume~35, pages 6679--6687, 2021.

\bibitem[Armstrong and Okun(2020)]{conventional_parkinson_disease}
Melissa~J. Armstrong and Michael~S. Okun.
\newblock Diagnosis and treatment of parkinson disease: A review.
\newblock \emph{JAMA}, 323\penalty0 (6):\penalty0 548--560, 02 2020.
\newblock ISSN 0098-7484.
\newblock \doi{10.1001/jama.2019.22360}.
\newblock URL \url{https://doi.org/10.1001/jama.2019.22360}.

\bibitem[Bakkialakshmi et~al.(2024)Bakkialakshmi, Arulalan, Chinnaraju, Ghosh, Rahat, and Saha]{Bakkialakshmi_Arulalan_Chinnaraju_Ghosh_Rahat_Saha_2024}
V~S Bakkialakshmi, V~Arulalan, Gowdham Chinnaraju, Hritwik Ghosh, Irfan~Sadiq Rahat, and Ankit Saha.
\newblock Exploring the potential of deep learning in the classification and early detection of parkinson’s disease.
\newblock \emph{EAI Endorsed Transactions on Pervasive Health and Technology}, 10, Mar. 2024.
\newblock \doi{10.4108/eetpht.10.5568}.
\newblock URL \url{https://publications.eai.eu/index.php/phat/article/view/5568}.

\bibitem[Bartl et~al.(2022)Bartl, Dakna, Schade, Wicke, Lang, Ebentheuer, Weber, Trenkwalder, and Mollenhauer]{bartl2022longitudinal}
Michael Bartl, Mohammed Dakna, Sebastian Schade, Tamara Wicke, Elisabeth Lang, Jens Ebentheuer, Sandrina Weber, Claudia Trenkwalder, and Brit Mollenhauer.
\newblock Longitudinal change and progression indicators using the movement disorder society-unified parkinson’s disease rating scale in two independent cohorts with early parkinson’s disease.
\newblock \emph{Journal of Parkinson’s Disease}, 12\penalty0 (1):\penalty0 437--452, 2022.

\bibitem[Biswas and Yadav(2023)]{biswat_et_al_CNN}
Barsha Biswas and Rajesh~Kumar Yadav.
\newblock A review of convolutional neural network-based approaches for disease detection in plants.
\newblock In \emph{2023 International Conference on Intelligent Data Communication Technologies and Internet of Things (IDCIoT)}, pages 514--518, 2023.
\newblock \doi{10.1109/IDCIoT56793.2023.10053428}.

\bibitem[Cabanillas-Carbonell and Zapata-Paulini(2025)]{CabanillasCarbonell2025Evaluation_zapata}
Michael Cabanillas-Carbonell and Jorge Zapata-Paulini.
\newblock Evaluation of machine learning models for the prediction of alzheimer's: In search of the best performance.
\newblock \emph{Brain, Behavior, and Immunity - Health}, 44:\penalty0 100957, January 2025.
\newblock ISSN 2666-3546.
\newblock \doi{10.1016/j.bbih.2025.100957}.
\newblock URL \url{https://doi.org/10.1016/j.bbih.2025.100957}.

\bibitem[Cheng(2024)]{cheng2024attention}
Liang Cheng.
\newblock Attention mechanism models for precision medicine, 2024.

\bibitem[Chintalapudi et~al.(2022)Chintalapudi, Battineni, Hossain, and Amenta]{Chintalapudi_Cascaded_DL}
Nandini Chintalapudi, Gopi Battineni, Md~Asif Hossain, and Francesco Amenta.
\newblock Cascaded deep learning frameworks in contribution to the detection of parkinson's disease.
\newblock \emph{Bioengineering}, 9\penalty0 (3):\penalty0 116, March 2022.
\newblock ISSN 2306-5354.
\newblock \doi{10.3390/bioengineering9030116}.
\newblock URL \url{https://doi.org/10.3390/bioengineering9030116}.

\bibitem[Dorsey and Bloem(2018)]{dorsey2018parkinson_call_to_action}
E~Ray Dorsey and Bastiaan~R Bloem.
\newblock The parkinson pandemic—a call to action.
\newblock \emph{JAMA neurology}, 75\penalty0 (1):\penalty0 9--10, 2018.

\bibitem[Foote et~al.(2025)Foote, de~Waal, Caiado, Samman, and Ukolov]{foote2025_comprehensive}
Avery Foote, Emma de~Waal, Frederico Caiado, Amr Samman, and Arkadiy Ukolov.
\newblock A comprehensive review of deep brain stimulation for parkinson’s disease: the history, current state of the art and future possibilities.
\newblock \emph{Medicine in Novel Technology and Devices}, page 100362, 2025.

\bibitem[Govindu and Palwe(2023)]{govindu2023early}
Aditi Govindu and Sushila Palwe.
\newblock Early detection of parkinson's disease using machine learning.
\newblock \emph{Procedia Computer Science}, 218:\penalty0 249--261, 2023.

\bibitem[Gutheil and Donsa(2022)]{gutheil2022saintens}
Julian Gutheil and Klaus Donsa.
\newblock Saintens: self-attention and intersample attention transformer for digital biomarker development using tabular healthcare real world data.
\newblock In \emph{dHealth 2022}, pages 212--220. IOS Press, 2022.

\bibitem[He et~al.(2024)He, Chen, Xu, Fortino, and Wang]{he2024early_ml_dl}
Tongyue He, Junxin Chen, Xu~Xu, Giancarlo Fortino, and Wei Wang.
\newblock Early detection of parkinson’s disease using deep neuroenhancenet with smartphone walking recordings.
\newblock \emph{IEEE Transactions on Neural Systems and Rehabilitation Engineering}, 2024.

\bibitem[Joseph(2023)]{Joseph03042023}
Claire~B. Joseph.
\newblock Parkinson disease.
\newblock \emph{Journal of Consumer Health on the Internet}, 27\penalty0 (2):\penalty0 221--224, 2023.
\newblock \doi{10.1080/15398285.2023.2212529}.
\newblock URL \url{https://doi.org/10.1080/15398285.2023.2212529}.

\bibitem[Khanna et~al.(2020)Khanna, Gambhir, and Gambhir]{khanna2020current_parkinson_research}
Ketna Khanna, Sapna Gambhir, and Mohit Gambhir.
\newblock Current challenges in detection of parkinson’s disease.
\newblock \emph{Journal of Critical Reviews}, 7\penalty0 (18):\penalty0 1461--1467, 2020.

\bibitem[Kilzheimer et~al.(2019)Kilzheimer, Hentrich, Burkhardt, and Schulze-Hentrich]{kilzheimer2019challenge_parkinson_predictive_power}
Alexander Kilzheimer, Thomas Hentrich, Simone Burkhardt, and Julia~M Schulze-Hentrich.
\newblock The challenge and opportunity to diagnose parkinson's disease in midlife.
\newblock \emph{Frontiers in Neurology}, 10:\penalty0 1328, 2019.

\bibitem[Kumar and Ujjwal(2024)]{kumar2024tabnet}
Tapan Kumar and RL~Ujjwal.
\newblock Tabnet unveils predictive insights: a deep learning approach for parkinson’s disease prognosis.
\newblock \emph{International Journal of System Assurance Engineering and Management}, pages 1--10, 2024.

\bibitem[Kurmi et~al.(2022)Kurmi, Biswas, Sen, Sinitca, Kaplun, and Sarkar]{kurmi2022_parkinson_ensemble}
Ankit Kurmi, Shreya Biswas, Shibaprasad Sen, Aleksandr Sinitca, Dmitrii Kaplun, and Ram Sarkar.
\newblock An ensemble of cnn models for parkinson’s disease detection using datscan images.
\newblock \emph{Diagnostics}, 12\penalty0 (5):\penalty0 1173, 2022.

\bibitem[Luo et~al.(2025)Luo, Qiao, Li, Wen, Zhang, and Li]{luo2025global}
Yuanrong Luo, Lichun Qiao, Miaoqian Li, Xinyue Wen, Wenbin Zhang, and Xianwen Li.
\newblock Global, regional, national epidemiology and trends of parkinson’s disease from 1990 to 2021: findings from the global burden of disease study 2021.
\newblock \emph{Frontiers in Aging Neuroscience}, 16:\penalty0 1498756, 2025.

\bibitem[Pahuja and Prasad(2022)]{pahuja2022deep}
Gunjan Pahuja and Bhanu Prasad.
\newblock Deep learning architectures for parkinson's disease detection by using multi-modal features.
\newblock \emph{Computers in Biology and Medicine}, 146:\penalty0 105610, 2022.

\bibitem[Pahwa and Lyons(2010)]{pahwa2010early_diagnosis}
Rajesh Pahwa and Kelly~E Lyons.
\newblock Early diagnosis of parkinson’s disease: recommendations from diagnostic clinical guidelines.
\newblock \emph{Am J Manag Care}, 16\penalty0 (4):\penalty0 94--99, 2010.

\bibitem[Prashanth and Roy(2018)]{prashanth2018early_parkinson_questionnaire}
R~Prashanth and Sumantra~Dutta Roy.
\newblock Early detection of parkinson’s disease through patient questionnaire and predictive modelling.
\newblock \emph{International journal of medical informatics}, 119:\penalty0 75--87, 2018.

\bibitem[Shastry(2023)]{K_ensemble_hard_voting}
K~Aditya Shastry.
\newblock An ensemble nearest neighbor boosting technique for prediction of parkinson’s disease.
\newblock \emph{Healthcare Analytics}, 3:\penalty0 100181, 2023.
\newblock ISSN 2772-4425.
\newblock \doi{https://doi.org/10.1016/j.health.2023.100181}.
\newblock URL \url{https://www.sciencedirect.com/science/article/pii/S2772442523000485}.

\bibitem[Singh et~al.(2025)Singh, Khare, Khare, and Kohli]{SINGH_et_al_2025}
Komal Singh, Manish Khare, Ashish Khare, and Neena Kohli.
\newblock Review on computational methods for the detection and classification of parkinson's disease.
\newblock \emph{Computers in Biology and Medicine}, 187:\penalty0 109767, 2025.
\newblock ISSN 0010-4825.
\newblock \doi{https://doi.org/10.1016/j.compbiomed.2025.109767}.
\newblock URL \url{https://www.sciencedirect.com/science/article/pii/S0010482525001179}.

\bibitem[Somepalli et~al.(2021)Somepalli, Goldblum, Schwarzschild, Bruss, and Goldstein]{saint}
Gowthami Somepalli, Micah Goldblum, Avi Schwarzschild, C.~Bayan Bruss, and Tom Goldstein.
\newblock Saint: Improved neural networks for tabular data via row attention and contrastive pre-training.
\newblock \emph{CoRR}, abs/2106.01342, 2021.
\newblock URL \url{http://dblp.uni-trier.de/db/journals/corr/corr2106.html#abs-2106-01342}.

\bibitem[Somvanshi et~al.(2024)Somvanshi, Das, Javed, Antariksa, and Hossain]{somvanshi2024survey}
Shriyank Somvanshi, Subasish Das, Syed~Aaqib Javed, Gian Antariksa, and Ahmed Hossain.
\newblock A survey on deep tabular learning.
\newblock \emph{arXiv preprint arXiv:2410.12034}, 2024.

\bibitem[ul~Haq et~al.(2022)ul~Haq, Li, Agbley, Mawuli, Ali, Nazir, and Din]{ul2022survey}
Amin ul~Haq, Jian~Ping Li, Bless Lord~Y Agbley, Cobbinah~Bernard Mawuli, Zafar Ali, Shah Nazir, and Salah~Ud Din.
\newblock A survey of deep learning techniques based parkinson’s disease recognition methods employing clinical data.
\newblock \emph{Expert Systems with Applications}, 208:\penalty0 118045, 2022.

\bibitem[Vaswani et~al.(2017)Vaswani, Shazeer, Parmar, Uszkoreit, Jones, Gomez, Kaiser, and Polosukhin]{vaswani2017attention}
Ashish Vaswani, Noam Shazeer, Niki Parmar, Jakob Uszkoreit, Llion Jones, Aidan~N Gomez, {\L}ukasz Kaiser, and Illia Polosukhin.
\newblock Attention is all you need.
\newblock \emph{Advances in neural information processing systems}, 30, 2017.

\bibitem[Wang et~al.(2020)Wang, Lee, Harrou, and Sun]{wang2020early}
Wu~Wang, Junho Lee, Fouzi Harrou, and Ying Sun.
\newblock Early detection of parkinson’s disease using deep learning and machine learning.
\newblock \emph{IEEE Access}, 8:\penalty0 147635--147646, 2020.

\bibitem[Xiao et~al.(2024)Xiao, Li, Feng, Wang, Zhu, and Chen]{xiao2024exploration}
Lingxi Xiao, Muqing Li, Yinqiu Feng, Meiqi Wang, Ziyi Zhu, and Zexi Chen.
\newblock Exploration of attention mechanism-enhanced deep learning models in the mining of medical textual data.
\newblock In \emph{2024 IEEE 2nd International Conference on Sensors, Electronics and Computer Engineering (ICSECE)}, pages 1251--1258. IEEE, 2024.

\bibitem[Zhang(2022)]{zhang2022mining}
Jing Zhang.
\newblock Mining imaging and clinical data with machine learning approaches for the diagnosis and early detection of parkinson’s disease.
\newblock \emph{npj Parkinson's Disease}, 8\penalty0 (1):\penalty0 13, 2022.

\bibitem[Zhang and Fan(2024)]{zhang2024memory}
Y~Zhang and Z~Fan.
\newblock Memory and attention in deep learning.
\newblock \emph{Academic Journal of Science and Technology}, 10\penalty0 (2):\penalty0 109--113, 2024.

\end{thebibliography}
\addcontentsline{toc}{chapter}{References}

\end{document}